\pdfoutput=1

\documentclass[11pt]{article}

\usepackage[final]{acl}

\usepackage{times}
\usepackage{latexsym}
\usepackage{amsmath}
\usepackage{multirow}
\usepackage{booktabs}
\usepackage{amssymb}
\usepackage{bbding}
\usepackage{pifont}
\usepackage{utfsym}

\usepackage[T1]{fontenc}

\usepackage[utf8]{inputenc}

\usepackage{microtype}

\usepackage{inconsolata}

\usepackage{graphicx}

\newcommand*\samethanks[1][\value{footnote}]{\footnotemark[#1]}

\title{RAVEN: Robust Advertisement Video Violation Temporal Grounding via Reinforcement Reasoning}

\author{Deyi Ji$^1$\thanks{The first two authors contribute equally to this work. We acknowledge Shaogang Tang for collaborating on data resources and application scenarios to validate and improve algorithm performance.} ~~Yuekui Yang$^{1,2}$\samethanks ~~Haiyang Wu$^1$ ~~Shaoping Ma$^2$ ~~Tianrun Chen$^3$ ~~Lanyun Zhu$^4$\thanks{Corresponding Author.} \\
  $^1$Tencent ~~~ $^2$Department of Computer Science and
Technology, Tsinghua University   \\ $^3$Zhejiang University  ~~~ 
$^4$Singapore University of Technology and Design  \\
  \texttt{deyiji@tencent.com,yuekuiyang@tencent.com,gavinwu@tencent.com, } \\
  \texttt{~~~msp@tsinghua.edu.cn,tianrun.chen@zju.edu.cn,lanyun\_zhu@mymail.sutd.edu.sg } 
  }

\begin{document}

\maketitle

\begin{abstract}
Advertisement (Ad) video violation detection is critical for ensuring platform compliance, but existing methods struggle with precise temporal grounding, noisy annotations, and limited generalization. We propose RAVEN, a novel framework that integrates curriculum reinforcement learning with multimodal large language models (MLLMs) to enhance reasoning and cognitive capabilities for violation detection. RAVEN employs a progressive training strategy, combining precisely and coarsely annotated data, and leverages Group Relative Policy Optimization (GRPO) to develop emergent reasoning abilities without explicit reasoning annotations. Multiple hierarchical sophisticated reward mechanism ensures precise temporal grounding and consistent category prediction. Experiments on industrial datasets and public benchmarks show that RAVEN achieves superior performances in violation category accuracy and temporal interval localization. We also design a pipeline to deploy the RAVEN on the online Ad services, and online A/B testing further validates its practical applicability, with significant improvements in precision and recall. RAVEN also demonstrates strong generalization, mitigating the catastrophic forgetting issue associated with supervised fine-tuning.

\end{abstract}

\section{Introduction}

In the modern digital landscape, advertisements play a pivotal role in sustaining the growth of internet platforms. To ensure compliance with local laws and regulations, promote sustainable development, and foster a user-friendly environment, platforms establish stringent guidelines to regulate the content uploaded by advertisers. Despite these efforts, violations of platform policies persist. Early approaches relied on small-scale models \cite{vit,resnet} to analyze and identify such violations, but these methods suffered from limited generalization capabilities. With the advent of large language models (LLMs) \cite{llava,qwen,zhu2024llafs,zhucpcf}, more advanced techniques have been increasingly adopted in practice to detect non-compliant content.

Among the various types of content, video advertisements present the most significant challenge for violation detection. In practice, it is not only necessary to predict the violation categories of a video but also to localize the specific sub-scenes corresponding to each category\cite{chen2024timemarker,urur,gu2024context,ji2024pptformer,ji2023guided}. A single video may contain multiple violation categories, each potentially associated with multiple temporal intervals. Existing methods typically follow a two-step process: (1) annotating each video with its violation categories and their corresponding temporal intervals, and (2) fine-tuning multimodal large language models (MLLMs) using supervised fine-tuning (SFT) techniques.

\begin{figure}[!ht]
      \centering
      \includegraphics[width=1\linewidth]{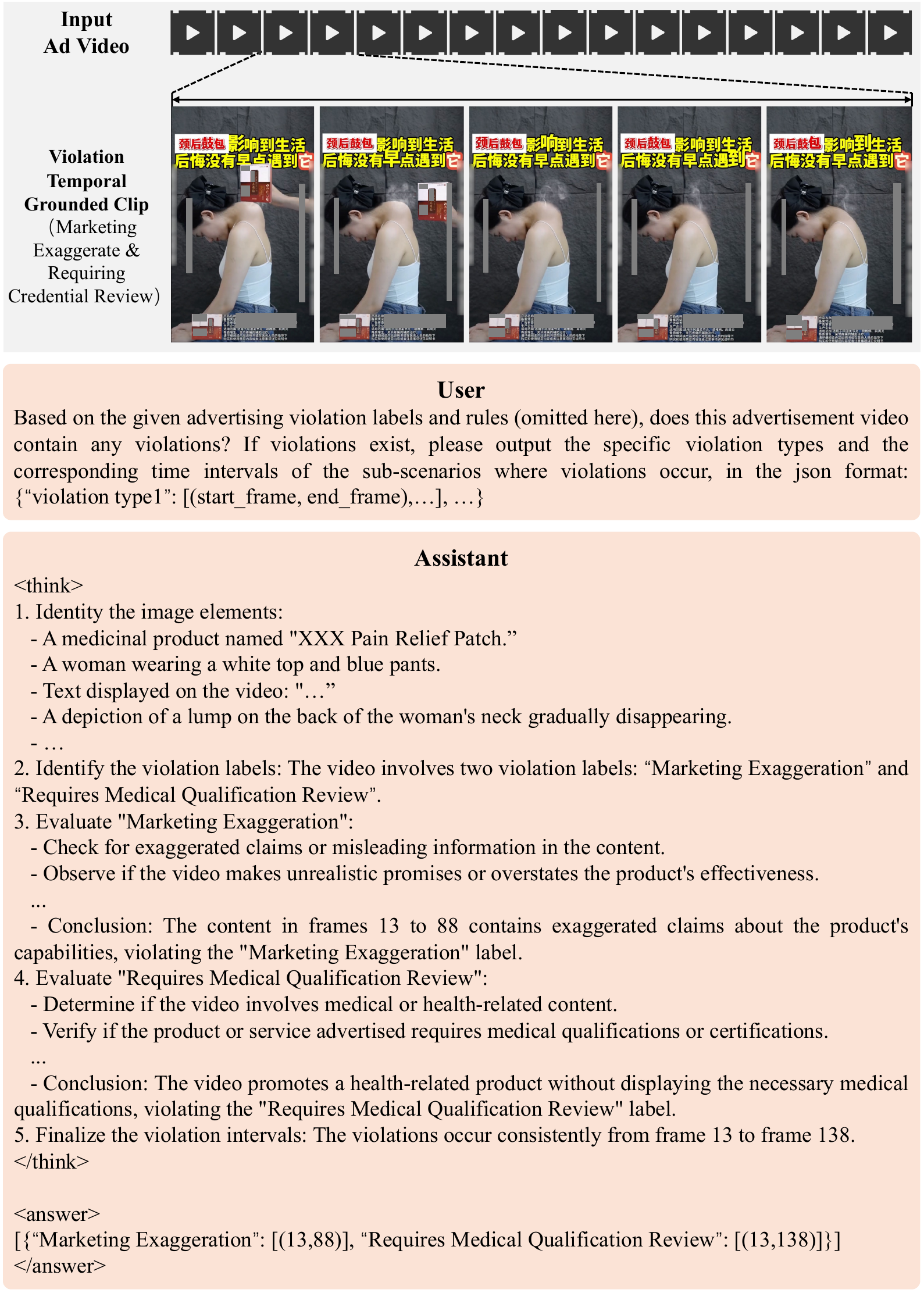}
      \caption{The example of RAVEN reasoning.}
      \label{fig_train_overview}
    \end{figure}

However, due to constraints in data volume, annotation costs, and the inherent difficulty of precise labeling, the annotated sub-scene intervals often contain natural errors or ambiguities. These inaccuracies can lead to unstable training or even misguided learning when using conventional SFT methods. As discussed in \cite{shao2024deepseekmath,liu2025seg,zhu2025popen}, SFT faces several limitations: its effectiveness notably diminishes in out-of-domain settings despite strong in-domain performance, it inherently causes catastrophic forgetting of general capabilities, and the absence of an explicit reasoning process limits its efficacy in complex scenarios. 
Recent research shows that pure reinforcement learning (RL) \cite{guo2025deepseek} fosters emergent reasoning during testing, underscoring the power of reward-driven optimization in boosting model reasoning. This method also tends to improve generalization, avoiding overfitting to specific datasets. 

Building on these insights, we introduce RAVEN, a novel framework aimed at advancing reasoning and cognitive skills for detecting violation scenes in videos. The name RAVEN, symbolizing ``keen insight'', reflects our aspiration for the system to detect violations with the sharpness of a raven. RAVEN is a structured reasoning MLLM that combines curriculum learning with hierarchical, multi-granular reinforcement. It employs GRPO (Group Relative Policy Optimization) \cite{shao2024deepseekmath,guo2025deepseek} and structured thinking, eliminating the need for explicitly annotated reasoning process data. Instead, it leverages the self-evolution potential of MLLMs to develop reasoning capabilities from scratch. A significant advantage of RAVEN is its ability to robustly train on large-scale, noisy, coarsely annotated industrial data, achieving superior violation detection performance while preserving the strong generalization capabilities of MLLMs. To achieve this, we develop hierarchical sophisticated rewards mechanism comprising multiple types of rewards: format rewards, which enforce constraints on the structure of the reasoning process and violation sub-scene outputs, and accuracy rewards, which include primary rewards (e.g., IoU Reward), auxiliary rewards (e.g., Boundary Alignment Reward), and regularization rewards (e.g., Category Consistency Reward). As illustrated in Figure 1, RAVEN exhibits emergent test-time reasoning abilities, enabling it to handle complex instructions by breaking them down into sequential analytical steps, thus achieving precise localization of violation intervals. RAVEN demonstrates exceptional performance on both in-domain and out-of-domain data, significantly outperforming models trained via SFT.

To validate RAVEN, we conduct extensive experiments from both offline and online testing perspectives, using both publicly available datasets and proprietary industrial data. The results show that the RAVEN-7B model exhibits strong test-time reasoning capabilities and achieves superior generalization performance compared to models of the same scale. Our contributions are threefold: (1) We propose RAVEN, the novel architecture specifically designed for localizing violation scenes in advertisement content. Through its innovative design, RAVEN exhibits emergent reasoning abilities. (2) RAVEN is a practical system tailored for real-world industrial applications. It demonstrates remarkable robustness when trained on large-scale, noisy, coarsely annotated data, while retaining strong generalization capabilities. (3) Extensive experiments on both offline and online testing, using public datasets and proprietary industrial data, demonstrate that the RAVEN-7B model achieves superior reasoning and generalization performance compared to models of the same scale.

\section{Related Work}

\subsection{Temporal Grounding in Videos}

Temporal grounding aims to localize specific events or actions within a video. Prior work has focused on supervised learning with precise annotations \cite{tall}. However, these methods struggle with noisy, coarsely annotated data, which is prevalent in industrial settings. Recent approaches like VSLNet \cite{zhang2020span} and 2D-TAN \cite{zhang2020learning} have improved localization accuracy but lack robust reasoning capabilities for complex tasks like violation detection.

\subsection{Multimodal Large Language Models}

Multimodal Large Language Models (MLLMs) \cite{yin2023survey,ji2024tree,xu2024large,toxvidlm,zhucpcf}, such as CLIP \cite{clip}, Flamingo \cite{flamingo}, and BLIP \cite{blip}, have demonstrated remarkable capabilities in understanding and reasoning across modalities on various tasks \cite{cot,dlpl,wei2022chain,urur,kojima2022large,sstkd,sstkd_pami,liu2024logic, zhu2024ibd}. These models excel in tasks like image-text retrieval and video captioning but are often limited by their reliance on supervised fine-tuning (SFT), which can lead to catastrophic forgetting and poor generalization. Recent efforts like LLaVA \cite{llava, llava_cot}, Qwen \cite{qwen,qwen_vl}  and Video-ChatGPT \cite{Video-ChatGPT} have explored integrating reasoning into MLLMs, but they remain underutilized in temporal grounding tasks.

\subsection{Reinforcement Learning for Video Understanding}

Reinforcement learning (RL) \cite{guo2025deepseek,rl,rlhf,zhu2024llafs,dpo,zhu2025llafs++,song2024preference,liustatistical} has been applied to video understanding tasks, such as action segmentation and event detection. Methods like SM-RL \cite{sm_rl,wang2019language} and RLPP \cite{rlpp} use RL to optimize temporal localization but are limited by their inability to handle multimodal inputs or perform complex reasoning. Curriculum reinforcement learning \cite{curriculum_rl,curriculum} has shown promise in improving RL’s robustness and generalization, but its application to temporal grounding remains unexplored.

\subsection{Advertisement Video Violation Detection}

Existing methods for advertisement video violation detection rely heavily on rule-based systems or supervised learning with precise annotations. These approaches are effective in controlled environments but fail to generalize to large-scale, noisy industrial datasets. Recent works \cite{wang2024mllm,lu2024towards} have explored using MLLMs for content moderation, but these methods lack the temporal grounding and reasoning capabilities required for precise violation detection. Our work bridges these gaps by introducing RAVEN, a curriculum reinforcement learning framework that integrates MLLMs with sophisticated reward mechanisms and structured reasoning for robust and precise advertisement video violation detection. By leveraging both precisely and coarsely annotated data, RAVEN addresses the limitations of existing methods and sets a new benchmark for temporal grounding in industrial applications.

\section{Methodology}

\subsection{Problem Overview}

Given an input video \( V \), a predefined list of violation labels \( T \), and a prompt \( P \), the Advertisement Video Violation Temporal Grounding task aims to output:  (1) The violation labels associated with the video.  
(2) The temporal intervals of the sub-scenes corresponding to each violation label.  Note that a single video may contain multiple violation labels, and each label may correspond to multiple sub-scenes. This requires the model to perform reasoning to accurately identify the most relevant frame fragments. Inspired by recent advancements in the reasoning capabilities of large models, we leverage this ability to develop a pipeline for reasoning-based violative sub-scene temporal grounding.  

We first employ reinforcement learning (RL) on a Multimodal Large Language Model (MLLM) to activate its reasoning ability, enabling it to generate a reasoning process and predict all violation categories \( \mathcal{C} = \{c_1, c_2, \dots, c_n\} \) and their corresponding accurate sub-scene locations \( \mathcal{X}_c = (t_c^l, t_c^r) \) for each category \( c \). Here, \( t_c^l \) and \( t_c^r \) denote the start and end times of the sub-scene, respectively.  

However, the manually annotated results \( \mathcal{Y}_c = (y_c^l, y_c^r) \) often deviate from the ground truth \( \mathcal{Z}_c = (z_c^l, z_c^r) \) due to annotation errors or ambiguities. To prevent supervised fine-tuning (SFT) from forcing the model to fit \( \mathcal{Y}_c \), which could lead to significant deviations from \( \mathcal{Z}_c \), we instead use RL for training. Additionally, to enhance the accuracy of the reasoning process, we follow DeepSeek \cite{deepseekmoe} and employ explicit structured thinking tags `<think>' for chained reasoning.

\subsection{Data Construction}

In real-world scenarios, for each advertisement video \( V \), when a violation is found, we annotate the precise violation category \( c \) and the corresponding temporal sub-interval \( \mathcal{Y}_c = (y_c^l, y_c^r) \) where the violation occurs. However, due to limitations in annotation resources, cost constraints, and inherent ambiguity in many videos, we can only maintain relatively accurate violation categories, while the annotated temporal intervals \( \mathcal{Y}_c \) often exhibit some degree of deviation from the ground truth \( \mathcal{Z}_c = (z_c^l, z_c^r) \). To address this, we organize the data based on a curriculum learning approach. Specifically, we select a subset of data with precisely annotated temporal intervals for the early stages of curriculum learning, while the remaining coarsely annotated data is used in the later stages. Additionally, it is important to note that for the reasoning training of RAVEN, we do NOT need to generate any offline reasoning data, meaning that RAVEN's reasoning does not require a cold-start training process.

\subsection{RAVEN Model}

We use Qwen2.5-VL \cite{qwen_vl} as the reasoning model \( F_{\text{reason}} \) in RAVEN. Although Qwen2.5-VL demonstrates some temporal grounding capabilities on public video understanding datasets, it struggles with accurate localization in real-world industrial applications. A straightforward approach would be to use precisely annotated temporal grounding data for SFT. However, acquiring large-scale, precisely annotated data is challenging and costly, especially for frame-level localization, which requires significant effort from annotators.  

Instead, we opt for coarse-grained annotations, which are faster and more cost-effective to produce. During the reinforcement learning stage, format rewards are employed to ensure the model generates structured outputs. This process can be formulated as:

\begin{equation}
\mathcal{C}, \mathcal{X} = F_{\text{reason}}(V, T, P),
\end{equation}

\noindent where \( \mathcal{C} \) represents the predicted violation categories, and \( \mathcal{X} \) denotes the corresponding temporal intervals.

Reasoning is a critical component in temporal grounding tasks. Inspired by DeepSeek-R1-Zero \cite{deepseekmoe}, we intentionally avoid using any explicit Chain-of-Thought (CoT) \cite{cot} data to teach RAVEN reasoning skills. Instead, we aim to activate its reasoning capabilities from scratch, enabling the model to autonomously generate a logical CoT before producing the final answer.  To achieve this, we design a structured user prompt and hierarchical sophisticated rewards  that guides the reasoning model to follow specific instructions. As shown in Figure 1, the user prompt instructs RAVEN to analyze and compare objects in the video, beginning by generating a reasoning process within `<think>' tags, followed by the final answer in a predefined format enclosed in `<answer>' tags.

\subsection{Reward Functions Design}

Reward functions play a pivotal role in RL, as they determine the optimization direction of the model. We manually design the following reward functions for RL:  

\subsubsection{Thinking Format Reward}

The reward mechanism is designed to facilitate a structured cognitive process within the model \cite{shao2024deepseekmath,guo2025deepseek}. Specifically, it directs the model to articulate its reasoning steps within the designated <think> and </think> tags, while the final output is to be presented between the <answer> and </answer> tags.

\subsubsection{Grounding Format Reward}

Our framework incorporates two levels of temporal grounding format rewards: soft and strict \cite{shao2024deepseekmath,guo2025deepseek}. The soft approach validates the format if temporal coordinates are included in the answer, regardless of their organization. The strict approach, however, mandates that the model follows the predefined structure exactly, utilizing specific keywords like "temporal start" and "temporal end" to achieve correctness.

\subsubsection{Temporal IoU Reward}

As the primary reward, the Temporal IoU Reward evaluates the overlap between the predicted sub-scene intervals \( \mathcal{X}_c \) and the annotated intervals \( \mathcal{Y}_c \). To maintain robustness against annotation noise, we binarize the IoU value using a threshold:  

\begin{equation}
R_{\text{IoU}} = \begin{cases} 
1 & \text{if IoU}(\mathcal{X}_c, \mathcal{Y}_c) > 0.5, \\
0 & \text{otherwise}.
\end{cases}
\end{equation}

\subsubsection{Temporal Boundary Alignment Reward}

Building on the IoU Reward, the Temporal Boundary Alignment Reward encourages the predicted interval boundaries \( (t_c^l, t_c^r) \) to align closely with the annotated boundaries \( (y_c^l, y_c^r) \). This reward is continuous and serves as an auxiliary reward with a smaller weight:  

\begin{equation}
R_{\text{Boundary}} = \exp\left(-\sigma^2 \left[(t_c^l - y_c^l)^2 + (t_c^r - y_c^r)^2\right]\right),
\end{equation}

\noindent where \( \sigma \) is a scaling factor.

\subsubsection{Violation Category Consistency Reward}

The Violation Category Consistency Reward ensures the predicted violation category \( c_p \) matches the annotated category \( c_g \). This reward is binary:  

\begin{equation}
R_{\text{Category}} = \begin{cases} 
1 & \text{if } c_p = c_g, \\
0 & \text{otherwise}.
\end{cases}
\end{equation}
\noindent where $c_p$ and $c_g$ indicates the prediction and groundtruth respectively.

\subsection{Curriculum Reasoning with Hierarchical Rewards}

RAVEN does not require a cold-start reasoning training process. We initiate training directly from the pre-trained Qwen2.5-VL model, utilizing the aforementioned rewards and applying the GRPO \cite{shao2024deepseekmath} algorithm in the subsequent curriculum reinforcement training process.

We utilize the Curriculum GRPO with hierarchical rewards, which leverages a combination of precisely annotated and coarsely annotated data, progressively refining the model’s ability to predict both the temporal intervals and the associated violation categories. The training process is divided into three stages, each designed to optimize specific aspects of the model’s performance.

\subsubsection{Stage 1: Training on Precisely Annotated Data} 

In the initial stage, the model is trained on a subset of data where the temporal intervals \( \mathcal{Y}_c = (y_c^l, y_c^r) \) are precisely annotated. The reward function for this stage is designed to ensure the model learns the overall position of the interval while also improving boundary precision and category consistency. The total reward \( R_{\text{Total}} \) is defined as:

\begin{equation}
R_{\text{Total}} = R_{\text{IoU}} + \alpha_1 \cdot R_{\text{Boundary}} + R_{\text{Category}},
\end{equation}

\noindent where \( R_{\text{IoU}} \) measures the overlap between the predicted interval \( \mathcal{X}_c \) and the annotated interval \( \mathcal{Y}_c \), binarized to ensure robustness against annotation noise.
\( R_{\text{Boundary}} \) encourages precise alignment of the predicted boundaries \( (t_c^l, t_c^r) \) with the annotated boundaries \( (y_c^l, y_c^r) \).
\begin{figure}[!ht]
      \centering
      \includegraphics[width=\linewidth]{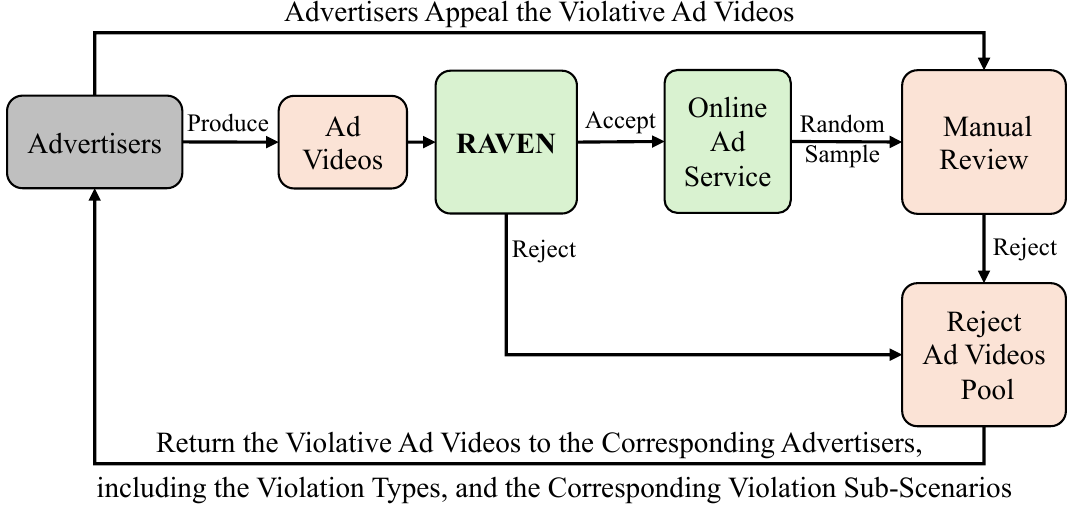}
      \caption{The deployment of RAVEN.}
      \label{fig_deployment}
    \end{figure}
\( R_{\text{Category}} \) ensures the predicted violation category \( c_p \) matches the annotated category \( c_g \). $\alpha_1$ is the reward weight. 
This stage focuses on establishing a strong foundation for interval prediction by prioritizing overall position (via \( R_{\text{IoU}} \)) while gradually refining boundary precision (via \( R_{\text{Boundary}} \)) and ensuring category consistency (via \( R_{\text{Category}} \)).

\begin{table*}[]
\centering
\scalebox{0.68}{\begin{tabular}{c|cc|cc|cc|cc|cc|cc}
\toprule
 \multirow{2}{*}{\textbf{Method}}                                                 & \multicolumn{2}{c|}{\begin{tabular}[c]{@{}c@{}}\textbf{Marketing} \\ \textbf{Exaggerate} \end{tabular}} & \multicolumn{2}{c|}{\begin{tabular}[c]{@{}c@{}}\textbf{Discomforting}\\\textbf{Content} \end{tabular}} & \multicolumn{2}{c|}{\begin{tabular}[c]{@{}c@{}}\textbf{Vulgar}\\ \textbf{Content} \end{tabular}} & \multicolumn{2}{c|}{\begin{tabular}[c]{@{}c@{}}\textbf{Requiring}\\ \textbf{Credential Review} \end{tabular}} & \multicolumn{2}{c|}{\begin{tabular}[c]{@{}c@{}}\textbf{Prohibited}\\ \textbf{Goods/Services} \end{tabular}}            & \multicolumn{2}{c}{\textbf{Average}} \\ \cmidrule{2-13} 
                          & {\color[HTML]{333333} Cate.(P/R)}        & {\color[HTML]{333333} Gro.} & {\color[HTML]{333333} Cate.(P/R)}     & {\color[HTML]{333333} Gro.}     & {\color[HTML]{333333} Cate.(P/R)}  & {\color[HTML]{333333} Gro.} & {\color[HTML]{333333} Cate.(P/R)}        & {\color[HTML]{333333} Gro.}        & {\color[HTML]{333333} Cate.(P/R)} & {\color[HTML]{333333} Gro.} & Cate.(P/R)        & Gro.       \\ \midrule

{\begin{tabular}[c]{@{}c@{}}Small\\ Models \end{tabular} } & 0.681/0.532  & - & 0.707/0.679  &  -  &  0.667/0.654  &  -   & 0.711/0.687 &  -  & 0.721/0.734    & - &  0.697/0.657     &  -       \\ \midrule
{\begin{tabular}[c]{@{}c@{}}LLaVA\\ -v1.5-SFT \end{tabular} } & 0.796/0.756  & 0.398 & 0.798/0.772  &  0.385  &  0.771/0.799  &  0.400    & 0.754/0.701 &  0.432  & 0.789/0.761    & 0.567 &  0.782/0.758     &  0.436       \\ \midrule
{\begin{tabular}[c]{@{}c@{}}Qwen2.5-VL\\ -7B-SFT \end{tabular} } & 0.832/0.787  & 0.424 & 0.821/0.798  &  0.402  &  0.800/0.810  &  0.411    & 0.773/0.702 &  0.461  & 0.797/0.771    & 0.580 &  0.805/0.774    &  0.456       \\ \midrule
{\textbf{RAVEN}} & 0.851/0.801  & 0.521 & 0.843/0.812  &  0.477  &  0.810/0.831  &  0.565    & 0.802/0.713 &  0.541  & 0.825/0.784    & 0.669 &  \textbf{0.826/0.788}     &  \textbf{0.555}       \\ \bottomrule
\end{tabular}}
\caption{Performance of Violation Category (Precision/Recall) and Violation Temporal Grounding (mIoU) on Industrial  Dataset. ``Cate." indicates ``Category'', and ``Gro." indicates ``Grounding''.}
\label{table_sota_industry}
\end{table*}

\begin{table}[]
\centering
\scalebox{0.8}{\begin{tabular}{c|cc}
\toprule
\multirow{2}{*}{\textbf{Method}}                                      & \multicolumn{2}{c}{\textbf{Average}}           \\ \cmidrule{2-3} 
                                                             & Cate. (P/R)          & Gro.           \\ \midrule
LLaVA-v1.5-SFT   & 0.509/0.501          & 0.370          \\ \midrule
Qwen2.5-VL-7B-SFT & 0.537/0.517          & 0.384          \\ \midrule
\textbf{RAVEN}                                               & \textbf{0.551/0.530} & \textbf{0.435} \\ \bottomrule
\end{tabular}}
\caption{Performance of Violation Category (Precision/Recall) and Violation Temporal Grounding (mIoU) on Public MultiHateClip Dataset.}
\label{study_hate}
\end{table}

\subsubsection{Stage 2: Training on the Large-Scale Coarsely Annotated Data}

In the second stage, the model is trained on data where the temporal intervals are coarsely annotated. Here, the reward function is simplified to focus on overall position and boundary alignment, as the imprecise nature of the annotations makes category consistency less reliable. The total reward \( R_{\text{Total}} \) is defined as:

\begin{equation}
R_{\text{Total}} = R_{\text{IoU}} + \alpha_2 \cdot R_{\text{Boundary}}.
\end{equation}

\noindent where $\alpha_2$ is the reward weight. By retaining \( R_{\text{IoU}} \) and \( R_{\text{Boundary}} \), the model learns to predict approximately correct intervals even with noisy annotations, while still improving boundary precision.

\subsubsection{Stage 3: Fine-Tuning on Full Dataset} 

In the final stage, the model is fine-tuned on the full dataset, combining both precisely and coarsely annotated data. The reward function is adjusted to balance overall position, boundary precision, and category consistency:

\begin{equation}
R_{\text{Total}} = \alpha_3 \cdot R_{\text{IoU}} + \alpha_4 \cdot R_{\text{Boundary}} + \alpha_5 \cdot R_{\text{Category}},
\end{equation}
\noindent where $\alpha_3$, $\alpha_4$, and $\alpha_5$ are the reward weights. This stage ensures the model achieves a robust balance between interval prediction and category identification, leveraging the strengths of both precise and coarse annotations.

\section{Deployment}

We design a pipeline to deploy the RAVEN on the online Ad services in Figure \ref{fig_deployment}, which include 3 parts: (1) RAVEN Review: It is the core of the entire pipeline, handling the primary review functions.
(2) Advertisers Appeal: It provides a channel for advertisers to appeal is they believe their ad is not violative.
(3) Manual Review: It is primarily applied in two scenarios. (a) Random Sampling Review: For Ads already published on the platform, random samples are reviewed to identify potential violations. This helps to: (i) address cases missed by the review model, and (ii) quickly detect new types of violations, providing decision-making references for subsequent model optimization. (b) Appeal Review: For cases that are appealed by advertisers, manual review provides the final decision.
(3) Model Iteration: Based on the continuously increasing volume and variety of online violation data, including (a) new types of violations, (b) more violation data, (c) difficult negative samples misidentified by the model, and (d) difficult positive samples missed by the model, we continuously iterate and optimize the RAVEN.

\section{Experiments and Results}

To comprehensively evaluate the performance of RAVEN, we conduct extensive experiments from both offline testing and online testing perspectives, utilizing both public dataset and practical industrial dataset. 

\subsection{Datasets} \label{sec_dataset}

To validate RAVEN’s performance in real-world industrial scenarios, we construct a dataset comprising approximately 38,000 training videos, which include both precisely annotated and coarsely annotated data, and 5,000 precisely annotated test videos. The use of a precisely annotated test set ensures reliability in evaluation. The annotations cover six major violation categories ( ``Discomforting Content", ``Marketing Exaggeration", ``Requiring Credential Review", ``Vulgar Content", ``Prohibited Goods/Services", and ``Normal") and the corresponding temporal intervals. The definitions of these major categories are inspired by both existing works \cite{wang2024mllm,wang2024multihateclip,lu2023facilitating} and the actual platform management rules. These major classes are further divided into multiple subcategories, forming a hierarchical and structured labeling system. In all experiments, we primarily focus on the major class labels to evaluate the model’s performance and robustness in high-level violation classification tasks.

\begin{table}[]
\centering
\scalebox{0.85}{\begin{tabular}{c|cc}
\toprule
\multirow{2}{*}{\textbf{Model}} & \multicolumn{2}{c}{\textbf{Online Sample Average}} \\ \cmidrule{2-3} 
                       & Cate.(P/R)               & Gro.                 \\ \midrule
Small Models          & 0.711/0.668         & -         \\ \midrule
Qwen2.5-VL-7B-SFT             & 0.800/0.787         & 0.478         \\ \midrule
RAVEN       & 0.821/0.803         &   0.563       \\ \bottomrule
\end{tabular}}
\caption{A/B Test on the Online Serving.}
\label{exp_online}
\end{table}

MultiHateClip \cite{wang2024multihateclip} is a publicly available dataset for hateful and offensive content detection on platforms like YouTube and Bilibili, featuring annotations for ``hateful", ``offensive", and ``normal" content.
Due to the unavailability of some videos, we conduct experiments on a downloadable subset of Bilibili, and manually annotate the temporal intervals.

\subsection{Offline Testing}

We compare RAVEN against several baseline models, including LLaVA-v1.5 \cite{llava}, Qwen2-VL-7B \cite{qwen_vl}, and Qwen2.5-VL-7B \cite{qwen_vl}, as well as their fine-tuned versions (SFT). The results in Table \ref{table_sota_industry} and Table \ref{study_hate} demonstrate that RAVEN significantly outperforms both the base pretrained models and the SFT models in ``violation category accuracy" and ``temporal grounding precision". Specifically, RAVEN achieves superior accuracy in sub-scene interval localization, highlighting the effectiveness of its curriculum reinforcement learning approach in enhancing the robustness of MLLMs.

\subsection{Online A/B Testing}

We conduct day-long online A/B testing on a practical business platform, allocating 20\% of the overall traffic for evaluation. RAVEN is compared against a small legacy model and Qwen2.5-VL-7B-SFT. The results in Table \ref{exp_online} show that RAVEN significantly improves violative video identification, achieving both higher precision and recall in category detection compared to the legacy model.  Additionally, RAVEN outperforms the Qwen2.5-VL-7B-SFT model by 8.5\% in temporal interval localization accuracy.

\subsection{Study on Generalization Capabilities}

As discussed in Section 1, SFT often leads to catastrophic forgetting of general capabilities, while RL enhances the generalization of MLLMs. To validate this claim, we conduct experiments on the Industrial dataset. Specifically, we train RAVEN on three in-domain categories (Discomforting Content, Marketing Exaggeration, Requiring Credential Review) and test it on the remaining two out-of-domain categories (Vulgar Content, Prohibited Goods/Services). The results in Table \ref{study_general} demonstrate that RAVEN, trained with RL, achieves higher accuracy and better generalization compared to the Qwen2.5-VL SFT model.

\begin{table}[]
\centering
\scalebox{0.8}{\begin{tabular}{c|cc}
\toprule
\multirow{2}{*}{\textbf{Model}}                                                                    & \multicolumn{2}{c}{\textbf{Average}} \\ \cmidrule{2-3} 
                                                                                          & Cate.(P/R)        & Gro.          \\ \midrule
Qwen2.5-VL-7B-SFT                                                                   & 0.805/0.774  & 0.456  \\ \midrule
RAVEN(w/o Structured Thinking) & 0.810/0.779  & 0.537  \\ \midrule
RAVEN  & 0.826/0.788  & 0.555  \\ \bottomrule
\end{tabular}}
\caption{Study on the Structured Thinking.}
\label{study_thinking}
\end{table}

\subsection{Study on Structured Thinking}

We further investigate the impact of reasoning training of structured thinking in RAVEN. Table 
\ref{study_thinking} shows that both w/o and w/ structured thinking outperform the SFT baseline, indicating that RL effectively boosts the model’s capabilities. However, RAVEN with structured thinking demonstrates even better performance, highlighting the importance of the reasoning process in handling complex video samples.

\subsection{Study on Reward Functions}

To validate the effectiveness of our reward function design, we conduct ablation studies on the format reward and temporal boundary alignment reward the on the Industrial dataset. The results in Table \ref{study_reward} demonstrate the effectiveness of the two reward functions.

\begin{table}[]
\centering
\scalebox{0.8}{\begin{tabular}{c|c|c}
\toprule
\textbf{Method} & \begin{tabular}[c]{@{}c@{}}In-Domain\\ (Average Gro.) \end{tabular}  & \begin{tabular}[c]{@{}c@{}}Out-of-Domain\\ (Average Gro.) \end{tabular} \\ \midrule
Qwen2.5-VL-7B-SFT   &   0.433   &   0.246         \\ \midrule
RAVEN  &    0.546      &  0.408          \\ \bottomrule
\end{tabular}}
\caption{Study on Generalization Capabilities.}
\label{study_general}
\end{table}

\begin{table}[]
\centering
\scalebox{0.6}{\begin{tabular}{c|c|c|c}
\toprule
\begin{tabular}[c]{@{}c@{}}Temporal Boundary \\ Alignment Reward\end{tabular} & \begin{tabular}[c]{@{}c@{}}Grounding Format\\  Reward\end{tabular} & \begin{tabular}[c]{@{}c@{}}Curriculum Reinforcemant \\ Learning\end{tabular} & Gro. 
 \\ \midrule
\usym{2613}                                  & strict                  & $\checkmark$            & 0.540 \\ \midrule
$\checkmark$           & soft                    & $\checkmark$            & 0.547 \\ \midrule
$\checkmark$             & strict                  &   \usym{2613}                                & 0.508 \\ \midrule
$\checkmark$             & strict                  & $\checkmark$            & 0.555 \\ \bottomrule
\end{tabular}}
\caption{Study on  Reward Functions and Curriculum Reinforcement Learning.}
\label{study_reward}
\end{table}

\subsection{Study on Curriculum Reinforcement Learning}

To evaluate the effectiveness of the curriculum reinforcement learning  strategy in RAVEN, we also conduct an ablation study on the Industrial dataset. As shown in Table \ref{study_reward}, when remove the progressive curriculum learning, the results shown in a significant drop in performance, with  temporal interval localization (mIoU) dropping by 4.7\%, highlighting the importance of leveraging multi-stage training for robust learning.

\section{Conclusion}

RAVEN is a novel framework for advertisement video violation detection, integrating curriculum reinforcement learning with multimodal large language models (MLLMs) to address challenges in temporal grounding and noisy annotations. Its progressive training strategy and hierarchical reward mechanism ensure precise localization and consistent category prediction. Experiments and online A/B testing demonstrate superior performance in accuracy, precision, and recall, while mitigating catastrophic forgetting. RAVEN establishes a promising methodological approach for practical violation detection, offering significant potential for advancing the field and addressing real-world challenges.

\section{Ethical Statement}

Our research adheres to ethical principles and prioritizes user rights. The dataset samples are for scientific analysis only and do not reflect the authors' views. All resources are intended for scientific research purposes only, contributing to the development of more secure and reliable digital platforms.

\bibliography{custom}

\end{document}